\documentclass[11pt,a4paper]{article}
\usepackage[hyperref]{emnlp2020}
\usepackage{times}
\usepackage{graphicx}
\usepackage{latexsym}
\usepackage{amsmath}
\usepackage{todonotes}

\newcommand\csr{\textsc{CsrQA}}

\newcommand\cs{\textsc{CsQA}}
\newcommand\crq{\textsc{CrQA}}
\usepackage{balance}

\newcommand\locnet{\textsc{SPNet}}

\usepackage{float}
\restylefloat{table}

\usepackage{microtype}

\aclfinalcopy 

\title{Joint Spatio-Textual Reasoning for Answering Tourism Questions}

  \author{Danish Contractor${^1}{^,}$${^2}$\thanks{This work was carried out as part of PhD research at IIT Delhi.The author is also a regular employee at IBM Research.} \quad Shashank Goel$^2$\thanks{Work carried out when the author was a student at IIT Delhi.}  \quad Mausam$^2$ \quad Parag Singla$^2$\\
$^1$IBM Research AI, New Delhi\qquad $^2$Indian Institute of Technology, New Delhi 
\qquad 
\\
\texttt{dcontrac@in.ibm.com, shashankgoel.iitd@gmail.com}, \\  \{\texttt{mausam,parags\}@cse.iitd.ac.in}}

\date{}
\begin{document}
\maketitle
\begin{abstract}


Our goal is to answer real-world tourism questions that seek Points-of-Interest (POI) recommendations. Such questions express various kinds of spatial and non-spatial constraints, necessitating a combination of textual and spatial reasoning. In response, we develop the first joint spatio-textual reasoning model, which combines geo-spatial knowledge with information in textual corpora to answer questions.
We first develop a modular spatial-reasoning network that uses geo-coordinates of location names mentioned in a question, and of candidate answer POIs, to reason over only spatial constraints. 
We then combine our spatial-reasoner with a textual reasoner in a joint model and present experiments on a real world POI recommendation task. We report substantial improvements over existing models without joint spatio-textual reasoning.

\end{abstract}

\section{Introduction} \label{sec:intro}
\begin{figure}
 {\footnotesize
 \center
   \includegraphics[scale=0.5]{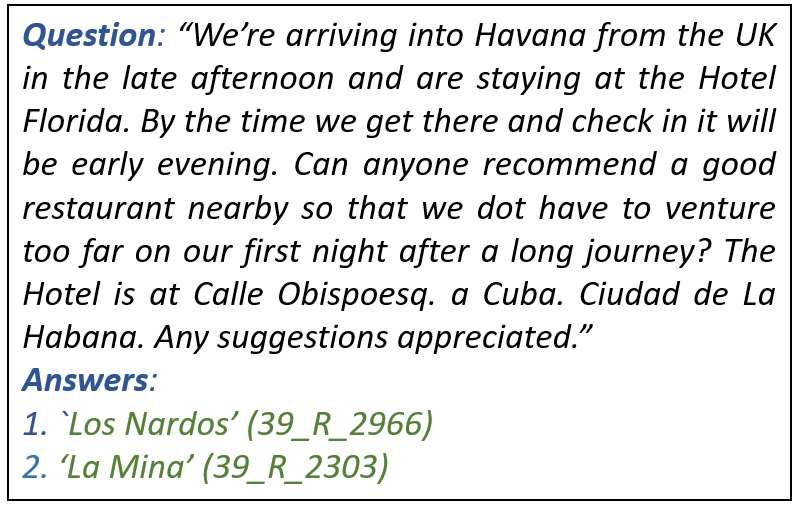}
   \caption{\small A sample POI recommendation question. The answers correspond to POI IDs of the form $<$city\_id $>$\_$<$POI type$>$\_$<$number$>$. The Tourism QA dataset has three classes of POIs - restaurants (R), attractions (A) and hotels (H).}
   \label{fig:queryExample}
 }
 \end{figure}

Users of travel forums often post questions seeking personalized recommendations for their travel needs. Consider the example in Figure \ref{fig:queryExample}, which shows a real-world\footnote{https://bit.ly/2zIxQpj} Points-of-Interest (POI) seeking question. Answering such a recommendation question is a challenging problem as, it not only requires reasoning over a text corpus describing potential restaurants (eg. reviews), but it also requires resolving spatial constraints (``near Hotel Florida'') over the physical location of a restaurant. In addition, the question is also under-specified and ambiguous (eg, ``dont have to venture too far'') making the spatial-inference task harder.

Recently, there has been work on QA models that fuse knowledge from multiple sources; for example, by combining data from knowledge bases with textual passages \cite{ExtKnowledge1,ExtKnowledge2}, 
or incorporating multi-modal data sources \cite{MultimodalQA2,MultimodalQA}. But, we do not know of systems that fuse geo-spatial knowledge with text. 
In addition, there exist several geo-spatial IR systems (eg, \cite{GikiCLEF,GeoAnalyticalQA1}), however, to the best of our knowledge, none of them perform joint-reasoning over geo-spatial and textual knowledge sources. 

In response, we present our joint spatio-textual QA model for returning answers to questions  that require textual as well as spatial reasoning. 
We first develop a modular spatial-reasoning network that uses geo-coordinates of location names mentioned in a question, and, of candidate answer entities, to reason over only spatial constraints. It learns to associate contextual {\em distance-weights} with each location-mention in the question 
-- these weights are combined with their respective spatial-distances from a candidate answer, to generate a `spatial relevance' score for that answer.

 We then combine the spatial-reasoner with a textual QA system to develop a joint spatio-textual QA model. We demonstrate the model using a recently introduced QA task, which contains tourism questions seeking POI (entity) answers \cite{TourQue}. It also contains a collection of entity reviews as knowledge source for answering these questions. We provide the geo-spatial knowledge for the task by mapping location-mentions in questions to their geographical coordinates using publicly available APIs. Similarly, candidate answer POIs are also mapped to their geographical coordinates, included as part of the dataset \cite{TourQue}. 
To the best of our knowledge, we are the first to develop a joint QA model that combines reasoning over external geo-spatial knowledge along with textual reasoning.

\noindent{\bf Contributions:}
Our paper makes the following contributions:

\noindent 1. We develop a spatial-reasoner that uses geo-coordinates of locations and POIs to reason over spatial constraints specified in a question. 


\noindent 2. We demonstrate, using a simple toy-dataset, that our spatial-reasoner is not only able to reason over  ``near'', ``'far'' constraints but is also able to determine location references that are not useful for reasoning (Eg: a location reference mentioning where a user last went on vacation). 

\noindent 3. We develop a spatio-textual QA model, which fuses spatial knowledge (geo-coordinates) with textual knowledge (POI reviews) using sub-networks designed for spatial and textual reasoning. 

\noindent 4. We demonstrate that our joint spatio-textual model performs significantly better than models employing only spatial- or textual-reasoning.
It also obtains state-of-the-art results on a real-world tourism questions dataset, with substantial improvement in answering location questions. 

\section{Related Work} \label{sec:related}
Our work is related to four broad areas of question answering and information retrieval:

\noindent {\bf Geographical Information Systems:} There is significant prior work on Geographical Information systems where standard IR models are augmented with spatial knowledge \cite{GIR2,GIR3}. Models have been developed to address challenges in adhoc-retrieval tasks with locative references \cite{GeoCLEF06,GEOCLEF08,GikiCLEF}. However, such models deal primarily with inference problems in toponyms (eg, ``Beijing is located in China"), location disambiguation and use of topographical classes (eg, ``Union lake is a water-body") etc. 
Methods for IR involving locative references use three strategies (i) a pipeline of filtering based on spatial information followed by text-based IR (ii) a pipeline of filtering based on text-based IR followed by ranking based on geo-spatial ranking or coverage, and (iii) a weighted or linear combination of two independent rankings \cite{ECIR2020}. Our work builds on the third strategy by jointly training a model with both geo-spatial and textual components. To the best of our knowledge, joint reasoning over text and geo-spatial data has not been investigated in geographical IR literature.

{\noindent \bf Geo-Spatial Querying:} There has been considerable work in research areas of geo-parsing (toponym discovery and disambiguation) \cite{toponym}, geo-spatial query processing over structured or RDF knowledge bases (KB) \cite{geospatial,GeoAnalyticalQA1}, geocoding and geo-tagging documents \cite{geocoding2,geotag,tweetloc} etc. However, such querying methods require KB \& task-specific annotations for training and are thus specialized in application and scope \cite{GeoAnalyticalQA1}.

\noindent{\bf Numerical Reasoning for Question Answering:} Spatial reasoning in our task is effectively a form of numerical reasoning over distances between location-mentions in a question and a candidate entity (POI). Recently introduced tasks such as DROP \cite{DROP} and QuaRTz \cite{QuARTZ} require reasoning that includes addition, subtraction, counting, etc. for answering reading comprehension style questions. Other tasks such as MathQA \cite{MathQA} and Math-SAT \cite{MathSAT} present high school and SAT-level algebraic word problems. 

Models developed for numerical reasoning tasks such as NAQANet \cite{DROP} and NumNet \cite{NumNet} reason over the explicit mentions of numerical quantities within a question or passage.  
In contrast, the questions in our task do not explicitly mention geographical coordinates, and also do not contain all the information required for numerical reasoning (since the distances need to be computed with respect to a candidate answer under consideration). Further, in contrast to algebraic word problems and numerical reasoning questions, answers in the POI-recommendation task are also heavily influenced by text-based reasoning on subjective POI-entity reviews.


\noindent{\bf Points-of-Interest (POI) Recommendation:} Existing models for POI recommendation typically rely on the presence of structured data, including geo-spatial coordinates. Queries may be structured or semi-structured and can consist of both spatial as well as textual arguments. Textual arguments are usually associated with the structured attributes or may serve as filters. Approaches include efficient indexing for `spatial' and `preference' features along with specialized data-structures as IR-Trees, \cite{VDLB09, ICDE2013, EDBT15,CIKM16}, methods based on Matrix Factorization \cite{matrixfac} for user-specific recommendations, click-through logs used for recommendations from search engines \cite{POI19} etc. 

Our work builds on the recently-released POI entity-recommendation QA task  \cite{TourQue,MSEQ}. Two approaches have been developed for this task: semantic parsing of unstructured user questions to query a semi-structured knowledge store \cite{MSEQ}, and an end-to-end trainable neural model operating over a corpus of unstructured reviews to represent POIs \cite{TourQue}.  Neither of these approaches explicitly reason on spatial constraints, even though the questions contain them. 



\section{Spatio-Textual Reasoning Network}
\label{sec:joint}
\begin{figure*}
\vspace{-2ex}
 {\footnotesize
 \center
   \includegraphics[scale=0.5]{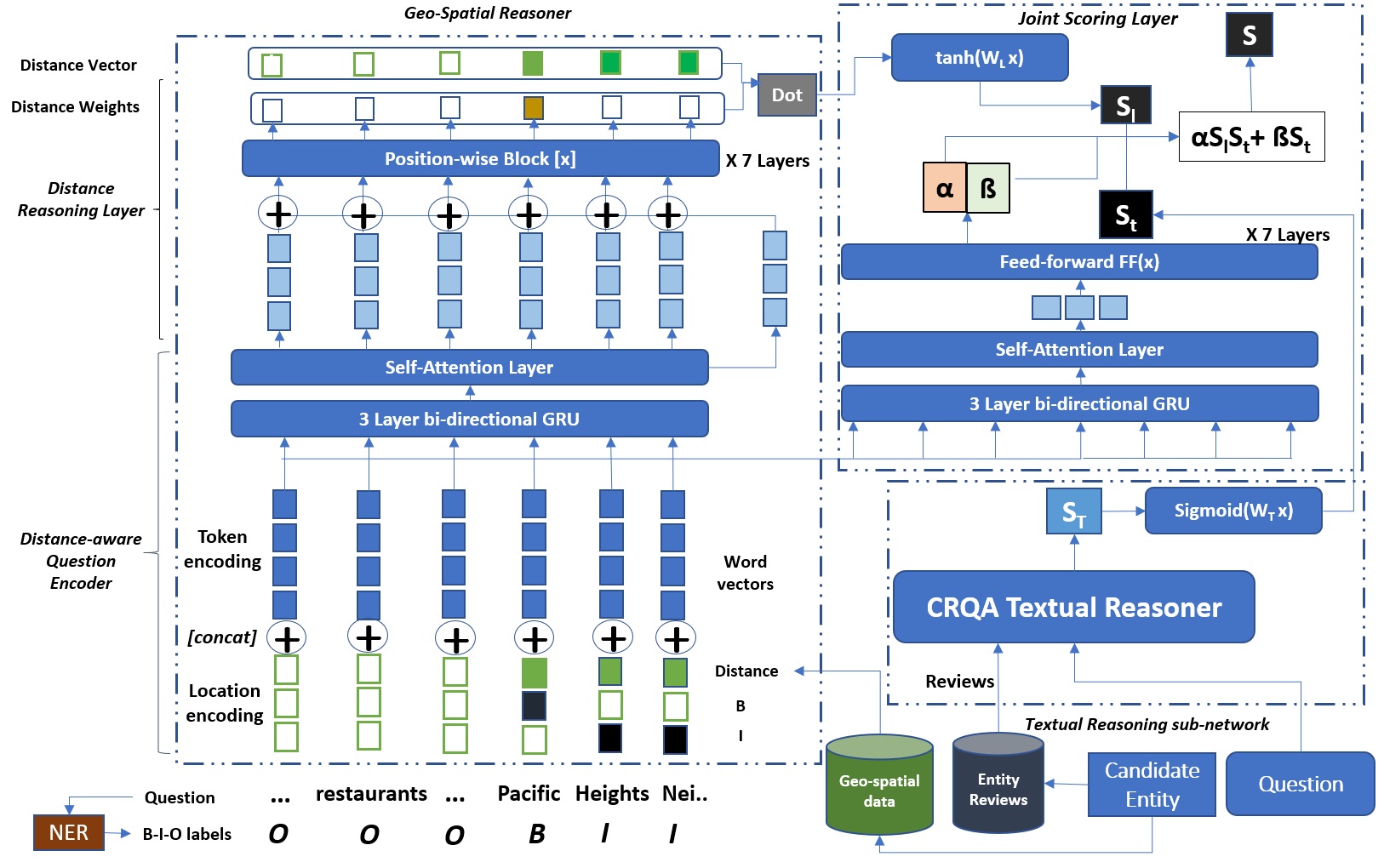}
   \caption{\small Spatio-Textual reasoning network consisting of (i) Geo-Spatial Reasoner (ii) Textual-Reasoning subnetwork (iii) Joint Scoring Layer}
   \label{fig:model}
 }
\vspace*{-3ex}
 \end{figure*}

The Spatio-Textual Reasoning Network (Figure \ref{fig:model}) consists of $3$ components: (i) Geo-Spatial Reasoner, (ii) Textual Reasoner, (iii) Joint Scoring Layer.  


\subsection{Geo-Spatial Reasoner} \label{sec:LRN}

Our geo-spatial reasoner consists of the following components: (1) {\bf Distance-aware Question Encoder} - to encode questions along with geo-spatial distances between location mentions (in the question) and a candidate entity, (2) {\bf Distance Reasoning layer} - to enable reasoning over geo-spatial distances with respect to the spatial constrains mentioned in the question, (3) {\bf Spatial Relevance Scorer} - to score and rank candidates for spatial-relevance. 

\noindent{\bf Distance-aware Question Encoder:} We generate question representations by using embedding representations of their constituent tokens along with embedding representations of their location-mentions. 
A question token can be represented by  traditional word-vector embeddings, or contextual embeddings such as BERT \cite{BERT}. Each token representation is further appended with a one-hot encoding representing Begin ({\bf B}), Intermediate ({\bf I}) or Other ({\bf O}) labels, indicating the presence of location tokens. The B-I labels help the model recognize a single continuous location-mention. In addition, we concatenate the distance\footnote{Manhattan Distance} of the candidate entity $c$, from a location-mention to each token-representation Thus, the question representations are distance-aware and candidate-dependent.  

Formally, let the token embedding representations in a question be given by $v_i$ $(v_0 \dots v_i \dots v_{m-1})$,where $m$ is the length of a question. Let the distance between the $k$th location-mention $lm_k$ and $c$ be denoted by $d_k$. Further, let $\phi(lm_k)$ be a function that returns the set of position indices occupied by location mention $lm_k$, i.e. it returns the set of position indices of question tokens that have been assigned the {\em B} or {I} label from the {\em B-I} encoding for location mention $lm_k$,  
 ($\phi(lm_k)$ $ \subset \{0,\dots,m-1\}$). 
We create an $m$-dimensional distance vector $ {\mathbf d'}$ where each element $d'_i$ of the vector is given by:
\begin{equation}
    d'_i= 
\begin{cases}
     d_k& \text{if i $\in$ {$\phi(lm_k$)}}\\
    0,              & \text{otherwise}
\end{cases}
\end{equation} 

Let the one-hot vector (two dimensional) of the {\em B-I} labels for the $i$th position be $g_i$. The input question embedding $t_i$, $(t_0 \dots t_i \dots t_{m-1})$ is then given by:
\begin{equation}
    t_i=concat[v_i,d'_i,g_i] 
    \label{eq:qe}
\end{equation}
We encode the question using a bi-directional GRU \cite{GRU} which results in output states $q_i$.
\noindent{\bf Distance-Reasoning Layer (DRL):} We first used a series of down-projecting feed-forward layers applied to the output state of the GRU, to generate the final score for each candidate, but we found this was not effective (Section \ref{sec:toy-eval}). 
We therefore include a component designed for distance-reasoning referred to as the `Distance Reasoning Layer' which uses the representations generated by the distance-aware question encoder.

A model could score candidate-entities for relevance if, for each location mentioned in the question, it is able to (i) learn whether a location-mention needs to be considered for answering, and (ii) learn {\em how} a location-mention needs to be used for answering.   
Our design of the DRL is motivated by this insight -- it learns a function which, for {\em each} location-mention $lm_k$, in the question, outputs a {\em distance-weight} $w_k$. Here, $w_k$ captures the contribution of the spatial-distance between $lm_k$ and the candidate entity $c$, under the constraints mentioned in the question. For instance, a question may include location-mentions that could be involved in simple `near' or `far' constraints or other complex constraints such as “within driving distance” or “within walking distance” etc. The DRL layer uses the distance-aware question encoding to understand the nature of the constraint being expressed, as well as, figure out how to compute distance-reasoning weights to express those constraints.

Let the output states of the question encoder be given by $q_0$ ..$q_i$.. $q_{m-1}$, where $m$ is the length of the question. 
To compute distance-weights, we use a series of position-wise feed-forward blocks \cite{Transformer} that consist of a linear layer with ReLU activation applied at each output position of the Question Encoder:
\begin{equation}
   q^l_i=Block_l(q^{l-1}_i)=max(0,A_l q^{l-1}_i+b_l)
\end{equation}
where $q^l_i$ is the output of the Block layer at layer $l$, $A_l$ is a weight matrix and $b_l$ the bias term. 

The initial block input uses the output state of the GRU ($q_i$) concatenated with the final hidden state ($q_L)$. Thus, the output $q^1_i$ from the application of the first block layer, corresponding position $i$ in the input is given by:
\begin{equation}
    q^1_i=Block_1(concat[q_i,q_{m-1}])
\end{equation}

The blocks apply the same linear transformations at each position but we vary the parameters across layers (see appendix). The final layer gives us a single dimension output for each position resulting in an $m$-dimensional vector ${\mathbf r}$ ($r_0...r_i..r_{m-1}$). 

Let ${\mathbf B}$ be an $m$-dimensional one hot-vector based on the position indices that have been assigned only the {\em B} label\footnote{An element of ${\mathbf B}$ is $1$ whenever it corresponds to a position index indicating the {\em start} of a location mention in a question.} from the {\em B-I} encoding used in the input layer.  
The distance-weight vector ${\mathbf w}$ for a question is given by: 

\begin{equation}
    {\mathbf w}= tanh(r \odot B) 
\end{equation}


We use the distance-weights for scoring, as described below.

\vspace{0.5ex}
\noindent{\bf Spatial Relevance Scorer:} 
The final score $S_L$ of a candidate $c$ is given by:
\begin{equation}
    S_L={\mathbf w} \dot {\mathbf d '}
\end{equation}
Note that since we concatenate the distance values along with token embeddings while encoding locations as part of the Question Encoder (Equation \ref{eq:qe}), it helps learn distance weights $w$ which are dependent on the distance value as well as the semantic information present in the question. Thus, the spatial relevance score is not just a simple linear combination of distances and makes the model representationally more powerful (see experiments in Section \ref{sec:joint-main-expts}). 
 We refer to the Geo-Spatial Reasoner as \locnet\ for brevity in the rest of the paper. 


\subsection{\bf Textual-Reasoning Sub-network:} We use the $\crq$ \cite{TourQue} model as our textual-reasoning sub-network. It consists of a Siamese-Encoder \cite{Siamese2} that uses question representations to attend over entity-review sentences and generate question-aware entity-embeddings. These entity embeddings are combined with question representations to generate an overall relevance score. For scalability, instead of using full review documents, the model uses a set of representative sentences from reviews after clustering them in USE-embedding space \cite{USE2018}. We follow \citeauthor{TourQue} and use k-means to cluster sentences in USE embedding space. We set $k$=10, and select 10 sentences per cluster, thus creating a $\le$ 100-sentence document to represent an entity.
In order to build a model that is capable of joint spatio-textual reasoning, our model learns question-specific combination weights that combine textual and spatial-reasoning scores.

\subsection{Joint Scoring Layer} Let the score generated by the textual-reasoner be $S_T$ and let the score generated by the spatial-reasoner be $S_L$. Let the rescaling weights for $S_T$ and $S_L$ be $w_T$ and $w_L$ respectively. 
Then, the overall score $S$ is given by: 
\begin{equation*}
\begin{aligned}
\alpha.\sigma(w_TS_T).\tanh(w_LS_L) + \beta.\sigma(w_TS_T),  
\end{aligned}
\end{equation*}
where $\sigma$ is the Sigmoid function and 
$\alpha$,$\beta$ are combination weights. The weights are computed by returning a two dimensional output (corresponding to each weight), after a series of feed-forward operations on the self-attended representation \cite{intra-attention}, of the question using the outputs of a Question Encoder with the same architecture as in \locnet (see appendix for hyperparameters). 
Note that the first term of scoring equation uses $S_L$ as a {\em selector}  
-- for questions where there are no locations mentioned, the spatial score of a question with no location-mentions will be $0$ (due to the equation for ${\mathbf w}$). 
This lets the model rely only on textual scores for these cases.



\noindent{\bf Training: } We train the joint model 
using max-margin loss, teaching the network to score  correct-answer entities higher than negative  samples. 

\section{Experiments}

We first present a detailed study of the spatial-reasoner using a simple artificially generated toy-dataset. This allows us to probe and study different aspects of spatial-reasoning in the absence of textual reasoning. We then present our experiments with the joint spatio-textual model using a real-world POI-recommendation QA dataset (Sec \ref{sec:joint-main-expts})

\subsection{Detailed Study: Geo-Spatial Reasoner}
We conduct this study on a simple toy-dataset generated using linguistically diverse templates specifying spatial constraints and location names chosen at random from a list of $200,000$ entities across $50$ cities.
\subsubsection{Artificial Toy-Dataset} \label{sec:toy}
\noindent{\bf Template Classes:} We create templates that can be broadly divided into three types of proximity queries based, on whether the correct answer entity is expected to be: 
(1) close to one or more locations (mentioned in the question), (2) far from one or more locations, (3) close to some and far from others (combination).
We create different templates for each category with linguistics variations. Figure \ref{fig:toyExample} shows a sample question from each category.  See appendix for more details, including the list of templates. 

\noindent{\bf Use of distractor-locations: }In order to make the task more reflective of real-world challenges we also randomly insert a {\em distractor} sentence that contains a location reference which does not need to be reasoned over (e.g the location ``Pinati'' in Question $2$ in Figure \ref{fig:toyExample}).  

\begin{figure}
 {\scriptsize
 \center
   \includegraphics[scale=0.50]{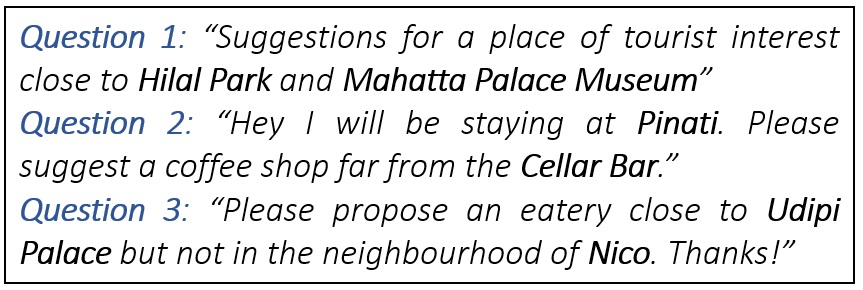}
   \caption{\small Sample questions from the Toy Dataset. The dataset has questions from three categories: (1) close to set X, (2) far from set X (3) Combination.}
   \label{fig:toyExample}
   \vspace{-2.5ex}
 }
\end{figure}

\begin{table*}
\scriptsize
\centering
\caption{Results of \locnet\ on the artificial spatial-questions dataset (t-test p-value $<10^{-33}$ for Acc@3)}
\resizebox{\textwidth}{!}{%
\begin{tabular}{|l|l|l|l|l|l|l|l|l|l|l|l|l|}
\hline
\multicolumn{1}{|c|}{\textbf{}} & \multicolumn{3}{c|}{\textbf{Close to set X}} & \multicolumn{3}{c|}{\textbf{Far from set X}} & \multicolumn{3}{c|}{\textbf{Combination}} & \multicolumn{3}{c|}{\textbf{Aggregate}} \\ \hline
\multicolumn{1}{|c|}{\textbf{Models}} & \multicolumn{1}{c|}{\textbf{Acc@3}} & \multicolumn{1}{c|}{\textbf{MRR}} & \multicolumn{1}{c|}{\textbf{$Dist_g$}} & \multicolumn{1}{c|}{\textbf{Acc@3}} & \multicolumn{1}{c|}{\textbf{MRR}} & \multicolumn{1}{c|}{\textbf{$Dist_g$}} & \multicolumn{1}{c|}{\textbf{Acc@3}} & \multicolumn{1}{c|}{\textbf{MRR}} & \multicolumn{1}{c|}{\textbf{$Dist_g$}} & \multicolumn{1}{c|}{\textbf{Acc@3}} & \multicolumn{1}{c|}{\textbf{MRR}} & \multicolumn{1}{c|}{\textbf{$Dist_g$}} \\ \hline
\locnet\ w/o DRL & 62.60 & 0.608 & 2.88 & 89.00 & 0.858 & 15.24 & 23.40 & 0.229 & 9.72 & 58.33 & 0.565 & 9.28 \\ \hline
\locnet\ & 90.20 & 0.873 & 0.860 & {\bf 98.00} & 0.975 & 13.88 & 52.80 & 0.486 & 3.90 & 80.33 & 0.778 & 6.21 \\ \hline \hline
BERT \locnet\ w/o DRL & 63.60 & 0.616 & 3.68 & 90.80 &0.881 & 15.32 & 26.80 & 0.242 & 12.96 & 60.40 & 0.579 & 10.65 \\ \hline
BERT \locnet\ & {\bf 91.40} & {\bf 0.896} & {\bf 0.78} & 97.80 & {\bf 0.978} & {\bf 13.87} & {\bf 59.20} & {\bf 0.551} & {\bf 3.02} & {\bf 82.80} & {\bf 0.808} & {\bf 5.89} \\ \hline
\end{tabular}%
}
\label{tab:toy-table}
\vspace{-1ex}
\end{table*}
\noindent{\bf Gold-entity generation:} The gold answer entity is uniquely determined for each question based on its template.  For example, consider a template T, ``{\em I am staying at \$A! Please suggest a hotel close to \$B but far from \$K.}" The score of a candidate entity $X$ is given by
$dist_T(X) = -(dist(X, B) - dist(X, K))$ (distances from {\em B} needs to be reduced, while distance from {\em K} needs to be higher). {\em A} is a distractor. The candidate with the $max$({$dist_T(X)$)} in the universe is chosen as the gold-answer entity for that question. We use the geo-coordinates of locations to compute the distance.

\noindent{\bf Dataset Statistics:} 
The train, dev and test sets consist of $6000$, $1500$ and $1500$ questions respectively generated using $48$ different templates, split equally across all $3$ template categories. Each question consists of location-names from only one city and thus the candidate search space for that question is restricted to that city. The average search space for each question is $1250$, varying between $10$-$16200$ across cities. The dataset includes questions containing distractor-locations ($52.33$\% of dataset)  distributed evenly across all template classes. 



\subsubsection{Results} \label{sec:toy-eval}
We study \locnet\ using the artificial dataset to answer the following questions: (1) What is the model performance across template classes? (2) How does the network compare with baseline models that do not use the DRL? 
(3) How well does the model deal with distractor-locations, i.e locations not relevant for the scoring task? For all experiments in this section we use perfectly tagged location-mentions. 

\noindent{\bf Metrics:} We study the performance of models using $Acc@N$ (N=3,5,30)\footnote{We report results with N=3 in the main paper. Please see appendix for full results.} which requires that any one of the top-N answers be correct, Mean Reciprocal Rank ($MRR$) and the average distance of the top-$3$ ranked answers from the gold-entity $Dist_g$. $Dist_g$ is helpful in quantifying the spatial goodness of the returned answers (lower is better). 

We use the following models in our experiments: (i) \locnet\  (ii) \locnet\ without DRL (iii) BERT-\locnet\ (iv) BERT-\locnet\ without DRL. 
Models without DRL use the final hidden states of the Question Encoder and a series of down-projecting feed-forward layers to generate the final score. 

\noindent{\bf Performance across template classes:} As can be seen in Table \ref{tab:toy-table}, all models perform the worst on the template class that contains a combination of both `close-to' and `far' constraints. Models based on \locnet\ perform exceeding well on the `Far' templates because the difference between the $dist_T(X)$ scores of the best and the second best candidate is almost always large enough for every model to easily separate them.

\noindent{\bf Importance of Distance-Reasoning Layer:} As can be seen in Table \ref{tab:toy-table} the performance of each configuration (with and without BERT) suffers a serious degradation in the absence of the DRL. Recall, that all models have access to spatial knowledge in their input layer via the question encoding. This indicates that the DRL is an important component required for reasoning on spatial constraints. To further assess whether our model is able to do distance reasoning, we computed the correlation between ranking-by-distances (appropriate ranking order for each template-class) and $\locnet$'s ranking on the toy-dataset. We found the rank correlation to be a high $0.97$ ($p<0.002$) suggesting that the model is able to use physical distance to compute the best answer.  

\begin{table}
\scriptsize
\centering
\caption{Performance of spatial-reasoning networks degrades in the presence of location-distractor sentences.}
\begin{tabular}{|l|l|l|l|l|l|l|}
\hline
\multicolumn{1}{|c|}{\textbf{}} & \multicolumn{2}{c|}{\textbf{Without Distractors}} & \multicolumn{2}{c|}{\textbf{With Distractors}} \\ \hline
\multicolumn{1}{|c|}{\textbf{Models}} & \multicolumn{1}{c|}{\textbf{Acc@3}} & \multicolumn{1}{c|}{\textbf{MRR}} & \multicolumn{1}{c|}{\textbf{Acc@3}} & \multicolumn{1}{c|}{\textbf{MRR}} \\ \hline
\locnet\ & 82.58 & 0.800 & 78.30 & 0.758  \\  \hline
BERT \locnet\ & {\bf 84.13} & {\bf 0.820} & {\bf 81.60} & {\bf 0.797} \\ \hline
\end{tabular}
\label{tab:toy-spurious-table}
\end{table}


\noindent{\bf Effect of distractor-locations:} We report results on two splits of the test set: Questions with and without distractor-locations. We report the aggregate performance over all template classes due to space constraints. As can be seen in Table \ref{tab:toy-spurious-table}, models suffer a degradation of performance in the presence of distractor-locations. We hypothesize that this is because the reasoning task becomes harder; models now need to also account for location-mentions that do not need to be reasoned over. 

\noindent{\bf Probing Study:}
\begin{figure}
{
    \center
    \includegraphics[scale=0.35]{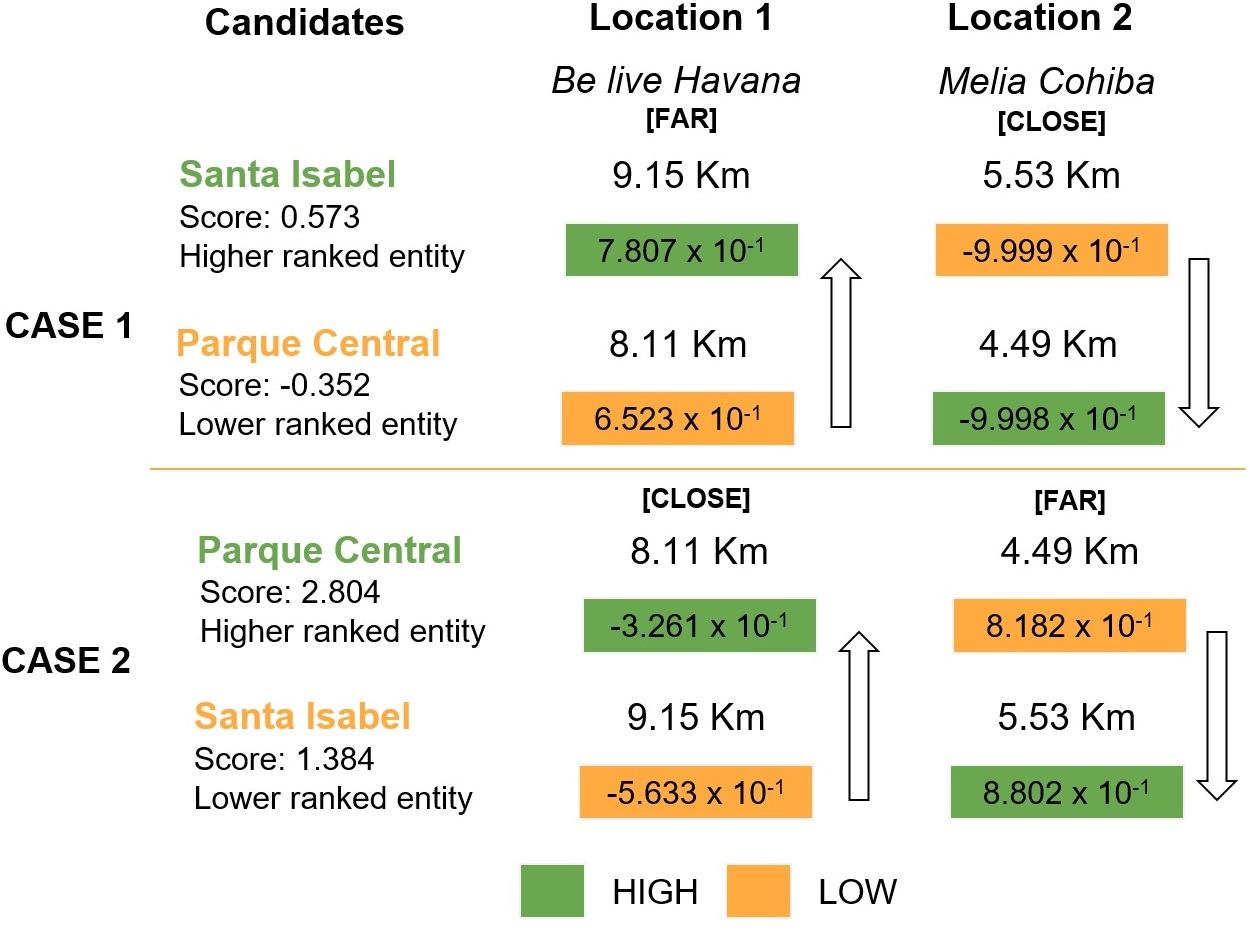}
    \scriptsize
    \caption{Probing study of the Distance Reasoning Layer (DRL) using the question: {\em ``I came from Tropicoco today. Any nice ideas for a coffee shop [\texttt{far from/close to}] `Be Live Havana' but [\texttt{close to/far from}] `Melia Cohiba'?''}}
    \label{fig:probe}
}
\vspace{-3ex}
\end{figure}
We conduct a probing study (Figure \ref{fig:probe}) on \locnet\ to get some insights into the reasoning process employed by the trained network. We use a question that has both `near' and `far' constraints (case 1) and then interchange the constraints (case 2). In both the cases we study the corresponding distance-weights assigned to the location-mentions with respect to two candidates ``Santa Isabel'' and ``Parque Central''.  Consider the first case; as can be seen, each candidate entity assigns a {\em higher} weight (column-wise comparison) as compared to the other candidate, on the distance property it is most likely to benefit from, with respect to the spatial-constraint. For example, when the spatial-constraint requires an answer to be {\em close} to ``Melia Cohiba'',  the candidate ``Parque Central'' assigns a higher weight to this location as compared to candidate ``Santa Isabel'', since ``Parque Central'' has a {\em smaller} distance value to this location. On the other hand, with respect to the ``{\em far}'' constraint, candidate ``Santa Isabel'' has a {\em larger} distance value from ``Be live Havana'' as compared to candidate ``Parque Central'', thus assigning a {\em higher} distance weight for this location-mention. 

When we interchange the constraints (Case $2$) we see the same pattern and the comparative weight trends (at each location-mention) invert due to inversion of spatial-constraints. This suggests, that DRL is learning to transform the inputs and generate weights based on the spatial constraint at hand. 

\noindent{\bf Effect of Candidate Space Size:} We analyzed the errors made by the \locnet\ model and we find that nearly 40\% of the errors were made in questions that have large ($>1000$) candidate spaces. Approximately $25\%$ of the test-set contains questions with large candidate spaces. 

\noindent{\bf Effect of the No. of Location-mentions:} The complexity of the spatial-reasoning task increases as the number of location-mentions (including distractor-locations) in the question increase. We find that \locnet\ makes no errors when spatial-reasoning involves only $1$ location-mention but, nearly 57\% of the errors are made in questions with $3$ location-mentions (See appendix).


\subsection{Spatio-Textual Reasoning Network} \label{sec:joint-main-expts}
For the joint model, we investigate the following research questions: (i) Does joint spatio-textual ranking result in improved performance over a model with only spatial-reasoning or only textual-reasoning? (ii)  How do pipelined baseline models that use spatial re-ranking perform on the task? (iii) Does distance-aware question encoding help in spatio-textual reasoning? (iv) Is the spatio-textual reasoning model more robust to distractor-locations as compared to baselines? (v) What kind of errors does the model make? 

\vspace{0.5ex}
\noindent{\bf Dataset:}
We use the recently released data set\footnote{https://github.com/dair-iitd/TourismQA} on Tourism Questions \cite{TourQue} that consists of over 47,000 real-world POI question-answer pairs along with a universe of nearly 200,000 candidate POIs; questions are long and complex, as presented in Figure \ref{fig:queryExample}, while the recommendations (answers) are represented by an ID corresponding to each POI. Each POI comes with a collection of reviews and meta-data that includes its geo-coordinates. The training set contains nearly $38,000$ QA-pairs and about $4,200$ QA-pairs each in  the validation and test sets. The average candidate space for each question is $5,300$. 

{\noindent \bf Task Challenges: } The task presents novel challenges of reasoning and scale; the nature of entity reviews (eg. inference on subjective language, sarcasm etc) makes methods such as BM25 \cite{BM25}, that are often used to prune the search space quickly in large scale QA tasks \cite{drqa,SearchQA}, ineffective. 
Thus, even simple BERT-based architectures or popular models such as BiDAF \cite{BIDAF} do not scale for the answering task in this dataset \cite{TourQue}.
 Thus, we use the non-BERT based \locnet\ subnetwork in the rest of the QA experiments\footnote{$\crq$ is also not based on BERT due to this reason.}.

\noindent{\bf Evaluation Challenges:} 
It is infeasible to construct a dataset of POI recommendation QA pairs, which has an exhaustively labeled answer-set for each question, since the candidate space is very large. Hence, this dataset suffers from the problem of false negatives, and $Acc@N$ metrics under-report system performance. Still, they are shown to be correlated with human relevance judgments \cite{TourQue}.  We therefore, use these metrics for all experiments, but additionally present a small human-study on the end-task, verifying the robustness of our results.



\noindent{\bf Location Tagging in Questions: }
In order to get mentions of locations in questions, we manually label a set of $425$ questions from the training set for location mentions. We then use a
BERT-based sequence tagger\footnote{github.com/codedecde/BiLSTM-CCM/tree/allennlp}
trained on this set to label locations.  The tagger has a macro-$F1$ of 88.03. This tagger tags all location mentions in a question without considering their utility for spatial-reasoning. 
It is possible that a question may contain {\bf only} distractor-locations, i.e., locations-mentions that do not need to be reasoned over the answering task.

Once the location-mentions are tagged, we remove the punctuations and stopwords from the tagged-location span. We then query the Bing Maps Location API\footnote{https://bit.ly/36Vazwo} using the location-mention along with the city (known from question meta-data) to get the geo-tags. To reduce noise in geo-tagging, we ignore the location-mention if the remaining text has a length of less than $4$ characters or is identified as a popular acronym, continent, country, city or state (lists from Wikipedia). We further reduce noise by ignoring a location mention: (1) if no results were found from BING, or (2) If the geo-tag is beyond 40km from the city center.  We found the location-mention geo-tagging precision on a small set of $83$ location-mentions to be 96\%. 
\begin{table}
\scriptsize
\centering
\caption{Dataset statistics: Questions with and without location-mentions across train, dev \& test sets from \cite{TourQue}.}
\begin{tabular}{|l|l|l|l|l|l|l|}
\hline
\multicolumn{1}{|c|}{\textbf{}} & \multicolumn{2}{c|}{\textbf{Location}} & \multicolumn{2}{c|}{\textbf{Non-location}} \\ \hline
\multicolumn{1}{|c|}{\textbf{Dataset}} & \multicolumn{1}{c|}{\textbf{Questions}} & \multicolumn{1}{c|}{\textbf{QA pairs}} & \multicolumn{1}{c|}{\textbf{Questions}} & \multicolumn{1}{c|}{\textbf{QA pairs}} \\ \hline
Train & 9,617 & 21,396 & 10,342 & 22,150 \\ \hline
Dev & 1,065 & 2,209 & 1,054 & 1,987 \\ \hline
Test & 1,086 & 2,198 & 1,087 & 2,144 \\ \hline
\end{tabular}
\label{tab:loc-stats-table}
\end{table}

We label all questions in the full dataset using this tagger, resulting in approximately 49.54\% of the QA pairs containing at least one location-mention (see Table \ref{tab:loc-stats-table}). In all our experiments, we use the Manhattan distance as our distance value, because it is generally closer to real-world driving/walking distance within a city, as opposed to straight-line distance. 

\begin{table}
\scriptsize
\center
\caption{Comparison of the joint Spatio-Textual model with baselines on questions that have location mentions (t-test p-value$<0.009$)
} 
\begin{tabular}{|c|l|l|l|l|l|}
\hline
\multicolumn{1}{|l|}{}                                          & \multicolumn{5}{|c|}{Location Questions}                                \\ \hline
\multicolumn{1}{|c|}{\bf Models}                                          & \textbf{Acc@3} & \textbf{Acc@5} & {\bf Acc@30} & \textbf{MRR} & $\mathbf{Dist_g}$ \\\hline
SD                                                            &           2.49    &       3.41       &  14.29   &    0.029     &       3.07      \\ \hline
\locnet\   &    1.47    &                                                2.11            &       8.47              & 0.019   &       2.97          \\ \hline \hline
CRQA                                                          &       14.83         &       21.27        &  50.65   &   0.143      &    3.41         \\ \hline
CRQA$\rightarrow$SD       &      13.73          &   19.26   &   50.65       &   0.125           &       {\bf 2.23}      \\ \hline
CRQA$\rightarrow$\locnet\     & 10.13               &  15.65   &    50.64       &  0.104            &     2.47        \\ \hline
\begin{tabular}[c]{@{}c@{}}Spatio-textual\\ CRQA\end{tabular} &    {\bf 18.32}        &       {\bf 25.69} &     {\bf 56.17}     &      {\bf 0.168}        &      2.62         \\ \hline
\end{tabular}
\label{tab:main-loc-results}
\end{table}
\subsubsection{Baselines}
Apart from the textual-reasoning model $\crq$ we also use the following baselines in our experiments:

\noindent{\bf Sort-by-distance (SD)}: Given a set of tagged-locations in a question and their geo-coordinates, rank candidate entities by the {\em minimum} distance from the set of tagged locations.

\noindent{\bf \locnet\ }: Use only the spatial-reasoning network for ranking candidate entities using their geo-coordinates. No textual-reasoning performed.

\noindent{\bf $\crq$ $\rightarrow$ SD:} Rank candidates using $\crq$ and then re-rank the top-30 answers using SD.

\noindent{\bf $\crq$ $\rightarrow$ \locnet\ }: Rank candidates using $\crq$ and then re-rank the top-30 answers using \locnet.

\begin{table*}
\scriptsize
\caption{Comparison of Spatio-Textual $\crq$ (with and without (w/o) distance-aware question encoding) and $\crq$ (t-test p-value $<0.03$ for $Acc@3$)} 
\begin{tabular}{|c|l|l|l|l|l|l|l|l|l|l|l|l|l|}
\hline
\multicolumn{1}{|l}{} & \multicolumn{5}{|c}{Location Questions} & \multicolumn{4}{|c}{Non-location Questions} & \multicolumn{4}{|c|}{Full Set} \\ \cline{2-14}
\multicolumn{1}{|c|}{Models} & \textbf{Acc@3} & \textbf{Acc@5} & \textbf{Acc@30} & \textbf{MRR} & \textbf{$Dist_g$} & \textbf{Acc@3} & \textbf{Acc@5} & \textbf{Acc@30} & \textbf{MRR} & \textbf{Acc@3} & \textbf{Acc@5} & \textbf{Acc@30} & \textbf{MRR} \\ \hline
CRQA & 14.83 & 21.27 & 50.65 & 0.143 & 3.41 & 18.95 & 26.22 & 54.37 & 0.177 & 16.89 & 23.75 & 52.51 & 0.159 \\ \hline

\begin{tabular}[c]{@{}c@{}}Spatio-Textual\\ CRQA\end{tabular} 
& {\bf 18.32} & {\bf 25.69} & {\bf 56.17} & {\bf 0.168} & {\bf 2.62 }& {\bf 20.42} & {\bf 26.77} & {\bf 56.49} & {\bf 0.18 }& {\bf 19.37} & {\bf 26.23} & {\bf 56.33} & {\bf 0.175} \\ \hline
 \begin{tabular}[c]{@{}c@{}}Spatio-textual\\ $\crq$ \\ ~(w/o distance-aware QE)\end{tabular} & 16.85 & 23.39  & {53.04} & 0.159  &2.84 & 20.06 & 26.86 & {56.49} & 0.185 & 18.45  & 25.13 & {54.76} & 0.172  \\ \hline
\end{tabular}
\label{tab:main-results}
\end{table*}

\noindent{\bf Training:} We pretrain \locnet\ on this dataset by allowing entities within a radius of $100$m from the actual gold-entity to be considered as gold (only for pretraining). To train the joint network we initialize model parameters learnt from component-wise pretraining of both \locnet\ as well as $\crq$. 

\subsubsection{Results} \label{sec:results-main}
We present our experiments on two slices of the test-set -- questions with tagged location-mentions (called {\em Location-Questions}) and those without any location mentions ({\em Non-Location Questions}). As can be seen in Table \ref{tab:main-loc-results} sorting-by-distance (SD) performs very poorly indicating that simple methods for ranking based on entity-distance do not work for such questions. Further, the poor performance of \locnet\ also indicates that the task cannot be solved just by reasoning on location data. 

In addition, pipelined re-ranking using SD or \locnet\ over the textual reasoning model decreases the average distance ($Dist_g$) from the gold-entity but does not result in improved performance in terms of answering (Acc@N) indicating the need for spatio-textual reasoning. Finally, from Tables \ref{tab:main-loc-results} \& \ref{tab:main-results} we note that the spatio-textual model performs better than its textual counterpart on the Location-Questions subset, while continuing to perform well on questions without location mentions. 

\noindent{\bf Effect of distance-aware question encoding:} In order to demonstrate the importance of distance-aware question encoding, we present an experiment where we remove the distance values from the input encoding. Thus, Equation \ref{eq:qe} changes to $t_i=concat[v_i,g_i]$. As Table \ref{tab:main-results} shows, the performance of the Spatio-Textual $\crq$ model in the absence of distance-aware encoding drops (last row), but it still performs better than the text-only $\crq$ model (first row). This indicates that the distance-aware question encoding helps learn better distance weights for spatio-textual reasoning.

\noindent{\bf Effect of distractor-locations:}
As mentioned earlier, we use a location-tagger that is oblivious to the reasoning task, to tag locations in the dataset.
We manually create a small set of $200$ questions, randomly selected from the test-set, but ensuring that half of it contains {\em at least} one non-distractor location mentioned in the question while the other half contains questions with {\em only} distractor-locations. 

\begin{table}[ht]
\scriptsize
\center
\caption{
Experiments on two subsets from the test-set: (i) Questions requiring Spatial-reasoning (ii) Questions with distractor-locations only.}
\begin{tabular}{|c|l|l|l|l|l|}
\hline
\multicolumn{1}{|l|}{}                                          & \multicolumn{5}{|c|}{Questions requiring Spatial-reasoning}                                \\ \hline
\multicolumn{1}{|c|}{\bf Models}                                          & \textbf{Acc@3} & \textbf{Acc@5} & {\bf Acc@30} & \textbf{MRR} & \textbf{$Dist_g$} \\\hline
SD                                                            &           5.00    &       7.00      &  22.00   &    0.053     &       2.10      \\ \hline
\locnet\   &    1.00    &                                                1.00           &       8.00            & 0.013   &       2.64          \\ \hline \hline
CRQA                                                          &       15.00         &       17.00        &  51.00   &   0.132      &    3.53         \\ \hline
CRQA$\rightarrow$SD       &      15.00          &   22.00   &   51.00       &   0.142           &       {\bf 1.963}      \\ \hline
CRQA$\rightarrow$\locnet\     & 16.00               &  23.00   &    51.00       &  0.134            &     2.41        \\ \hline
\begin{tabular}[c]{@{}c@{}}Spatio-textual\\ CRQA\end{tabular} &    {\bf 22.00}        &       {\bf 28.00} &     {\bf 54.00}     &      {\bf 0.182}        &      2.62         \\ \hline
\multicolumn{1}{|l|}{}                                          & \multicolumn{5}{|c|}{{Questions with distractor-locations only}}                                \\ \hline
SD                                                            &           2.00    &       3.00       &  17.00   &    0.025     &       4.12      \\ \hline
\locnet\   &    1.00    &                                                2.00           &       9.00              & 0.016   &       4.14          \\ \hline \hline
CRQA                                                          &       19.00         &       26.00        &  51.00   &   0.162      &    3.62         \\ \hline
CRQA$\rightarrow$SD       &      13.00          &   17.00  &  51.005       &   0.108          &       { 3.26}      \\ \hline
CRQA$\rightarrow$\locnet\     & 13.00               &  17.00   &    51.00       &  0.113            &     {\bf 3.24}        \\  \hline
\begin{tabular}[c]{@{}c@{}}Spatio-textual\\ CRQA\end{tabular} &    {\bf 20.00}        &       {\bf 28.00} &     {\bf 53.00}     &      {\bf 0.187}        &      3.50         \\ \hline
\end{tabular}
\label{tab:main-loc-clean-results}
\end{table}

As can be seen from Table \ref{tab:main-loc-clean-results}, 
all models including the spatio-textual model deteriorate in performance if a question only contains distractors; the spatio-textual model however, suffers a less significant drop in performance. 


\noindent{\bf Qualitative Study:}
\begin{table}
\scriptsize
\centering
\caption{Spatio-Textual $\crq$: Classification of Errors}
\begin{tabular}{|c|c|}
\hline
\textbf{Error Type} & \textbf{Percentage} \\ \hline
Textual Reasoning Error & 37.9\% \\ \hline
Far from the required location & 22.3\% \\ \hline
Influenced by Distractor & 12.6 \% \\ \hline
Not in requested Neighbourhood & 10.7 \% \\ \hline
Location Tagger Error & 5.8 \% \\ \hline
RepeatedLocation Names & 4.9 \% \\ \hline
Error in Geo-Spatial Data & 2.9 \% \\ \hline
Invalid Question & 2.9 \% \\ \hline
\end{tabular}
\label{tab:main-error-table}
\end{table}
We randomly selected $150$ QA pairs with location-mentions from the test-set, to conduct a qualitative error analysis of Spatio-textual $\crq$ (Table \ref{tab:main-error-table}). We find that nearly 37\% of the errors can be traced to the textual-reasoner, 22\% of the errors were due to a `near' constraint not being satisfied, while about 13\%  of the errors were due to the model reasoning on distractor-locations. Lastly 8\% of the errors were due to errors made by the location-tagger and incorrect geo-spatial data.

\noindent{\bf Effect of Candidate Search Space:}
\begin{table}
\scriptsize
\caption{Comparison with current state-of-the-art $\csr$ on (i) Location Questions (ii) All data}
\center
\begin{tabular}{|c|l|l|l|l|l|}
\hline
\multicolumn{1}{|l|}{}                                          & \multicolumn{5}{|c|}{Location Questions}                                \\ \hline
\multicolumn{1}{|c|}{\bf Models}                                          & \textbf{Acc@3} & \textbf{Acc@5} & {\bf Acc@30} & \textbf{MRR} & \textbf{$Dist_g$} \\\hline
$\csr$ & 19.89 & 26.43 & {\bf 51.47} & 0.168 & 2.70 \\ \hline
\begin{tabular}[c]{@{}c@{}}Spatio-textual\\ CSRQA\end{tabular} 
& {\bf 21.36} & {\bf 28.36} & {\bf 51.47} & {\bf 0.183} & 2.27 \\ \hline
\multicolumn{1}{|l|}{}                                          & \multicolumn{5}{|c|}{All Questions}                                \\ \hline

$\csr$ & 21.45 & 28.21 & {\bf 52.65} & 0.186 & 2.47 \\ \hline
\begin{tabular}[c]{@{}c@{}}Spatio-textual\\ CSRQA\end{tabular} 
& {\bf 22.41} & {\bf 28.99} & {\bf 52.65} & {\bf 0.193} & {\bf 2.32}\\ \hline
\end{tabular}

\label{tab:csr-loc-results}
\end{table}
Past work \cite{TourQue} has improved overall task performance by employing a neural IR method to reduce the search space \cite{DUET2}, and then using the $\crq$ textual-reasoner to re-rank only the top 30 selected candidates (pipeline referred to as $\csr$). 
In line with their work, we create a spatio-textual counterpart to $\csr$, by using spatio-textual reasoning in re-rank step.  
We find that this final model results in a $1$ pt (Acc@3) improvement overall (see Table \ref{tab:csr-loc-results}), and a $1.5$ pt improvement on location questions (Acc@3), establishing a new state of the art on the task. We note that, because the IR selector is incapable of spatial-reasoning, it possibly reduces the gains made by the spatio-textual re-ranking. An interesting direction of future work could be to augment general purpose neural IR methods with such spatial-reasoning. 

\noindent {\bf Effect of False Negatives:} To supplement the automatic evaluation, we additionally conducted a blind human-study using the top-ranked $\csr$ and spatio-textual $\csr$  models on another subset of $100$ questions from the test-set. Two human evaluators ($\kappa$=0.81) were presented the top-3 answers from both models in random order and were asked to mark each answer for relevance. 
\begin{table}
\scriptsize
\centering
\caption{Acc@3 results on a blind-human study using $100$ randomly selected questions from the test-set} 
\begin{tabular}{|l|l|l|l|l|l|l|}
\hline
\multicolumn{1}{|c|}{\textbf{}} & \multicolumn{2}{c|}{\textbf{Automated evaluation}} & \multicolumn{2}{c|}{\textbf{Human evaluation}} \\ \hline
\multicolumn{1}{|c|}{\textbf{}} & \multicolumn{1}{c|}{\textbf{Location}} & \multicolumn{1}{c|}{\textbf{Non-location}} & \multicolumn{1}{c|}{\textbf{Location}} & \multicolumn{1}{c|}{\textbf{Non-location}} \\ \hline
$\csr$ & 28.00 & 36.00 & 64.00 & 70.00  \\  \hline
\begin{tabular}[c]{@{}c@{}}Spatio-textual\\ CSRQA\end{tabular} 
& 32.00 & 32.00 & 84.00 & 72.00 \\ \hline
\end{tabular}
\label{tab:main-humaneval-table}
\end{table}
As Table \ref{tab:main-humaneval-table} shows, the manual annotation resulted in Acc@3 for $\csr$ and spatio-textual $\csr$ at a much higher, $67$\% and $78$\% respectively. On the subset of location questions, the accuracy numbers are $64\%$ and $84\%$.
This underscores the value of joint spatio-textual reasoning for the task, and signifies a substantial improvement in the overall QA performance.

\section{Conclusion} \label{sec:conclusion}
Our paper presents the first joint spatio-textual QA model that combines spatial and textual reasoning. 
Experiments on an artificially constructed (spatial-only) toy QA dataset show that our spatial reasoner effectively trains to satisfy spatial constraints. 
We also presented detailed experiments on the recently released POI recommendation task for tourism questions. Comparing against textual only and spatial only QA models, the joint model obtains significant improvements.  Our final model establishes a new state of the art on the task. In future work, we would like to also support reasoning on questions that require directional or topographical inference (eg.``north of X'', ``on the river beach'').

\section{Acknowledgements}
 We would like to thank Krunal Shah, Gaurav Pandey, Biswesh Mohapatra, Azalenah Shah for their helpful suggestions to improve early versions of this paper. We would also like to acknowledge the IBM Research India PhD program that enables the first author to pursue the PhD at IIT Delhi. This work is supported by an IBM AI Horizons Network grant, IBM SUR awards, Visvesvaraya faculty awards by Govt. of India to both Mausam and Parag as well as grants by Google,  Bloomberg  and 1MG  to Mausam.

\bibliographystyle{acl_natbib}
\bibliography{emnlp2020}

\appendix

\appendix

\section{Appendix}
This appendix is organized as follows. 
\begin{itemize}
    \item Section \ref{sec:artificial} provides more details about the Toy Dataset and supplementary experimental information that includes additional tables referred to in the main paper on Spatial Reasoning.
    \item Section \ref{sec:loctagger} includes more results of the Location Tagger used in the end-task. 
     \item Section \ref{sec:spatio-textual-expts} contains supplementary experiments on Spatio-Textual Reasoning. 
    \item Section \ref{sec:model-params} gives details about the model hyper-parameters.
   
\end{itemize}
\subsection{Toy Dataset} \label{sec:artificial}

We create a simple toy-dataset that is generated using linguistically diverse templates specifying spatial constraints and locations chosen at random from across 200,000 entities. These entities were sourced from the recently released Points-of-Interest (POI) recommendation task \cite{TourQue}. Each POI entity is labeled with its geo-coordinates apart from other meta-data such as its address, timings, etc. Further, each entity in a city has a specific type viz. Restaurant(R), Attraction(A) or Hotel(H). Table \ref{tab:templates-table} shows the list of templates used for generating the dataset. These templates have been to make the toy-dataset reflective of real-world challenges. For instance, templates \#41-\#48 include the possibility of injecting {\em distractor locations}. To generate questions,  \$LOCATION and \$ENTITY values are updated by randomly selecting values from the POI-set for each entity as described in the next section.

\subsubsection{Dataset Generation}

To generate a question, a city \textit{c}, type \textit{t} and a template \textit{T} are chosen at random. The "ENTITY" token in each template is replaced by a randomly chosen \textit{metonym} of the type \textit{t}. Table  \ref{tab:toy-metonyms-table} shows the list of metonyms for each type. Each instance of the "LOCATION" token is replaced by a randomly chosen entity from the city \textit{c} and type \textit{t}. The candidate set consists of the entities from the city \textit{c} and type \textit{t}. The entities used as location mentions are sampled without replacement and removed from the candidate set. 

The gold answer entity is uniquely determined for each question based on its template.  For example, consider a template T, ``{\em I am staying at \$A! Please suggest a hotel close to \$B but far from \$C.}" The score of a candidate entity $X$ is given by
$dist_T(X) = -(dist(X, B) - dist(X, C))$ (distances from {\em B} needs to be reduced, while distance from {\em C} needs to be higher). {\em A} is a distractor. The candidate with the $max$({$dist_T(X)$)} in the universe is chosen as the gold entity for that question. 

Each question further consists of 500 negative samples (35\% hard, 65\% soft). The negative samples are generated as a part of the gold generation process. A hard negative sample has a $dist_T(X)$ value closer to the gold as compared to a soft negative sample. We release the samples used for training along with the dataset for reproducibility.

\setlength\doublerulesep{0.2cm} 
\begin{table*}
\scriptsize
\centering
\caption{Templates used for generating the Toy-dataset} 
\begin{tabular}{|l|l|}
\hline
\multicolumn{1}{|c|}{\textbf{Id}} & \multicolumn{1}{c|}{\textbf{Description}} \\ \hline
1 & Do you have any recommendations of \textit{ENTITY} near the \textit{LOCATION}? \\ \hline
2 & Does anyone have ideas on \textit{ENTITY} close to \textit{LOCATION}? Thank you! \\ \hline
3 & Hello! Could anyone please suggest \textit{ENTITY} in the neighborhood of \textit{LOCATION}? \\ \hline
4 & Good Morning! Can someone please propose \textit{ENTITY} not very far from \textit{LOCATION}? \\ \hline
5 & Suggestions for \textit{ENTITY} close to both \textit{LOCATION} and \textit{LOCATION}? \\ \hline
6 & Some good ideas of \textit{ENTITY} between \textit{LOCATION} and \textit{LOCATION}? Thanks much! \\ \hline
7 & Please advise \textit{ENTITY} close to \textit{LOCATION} and not very far off the \textit{LOCATION}. \\ \hline
8 & Any ideas for \textit{ENTITY} near \textit{LOCATION} and also close to \textit{LOCATION} would be welcomed? \\ \hline \hline
9 & I once lived around \textit{LOCATION}. Does anyone have ideas of \textit{ENTITY} close to the \textit{LOCATION}? Thanks! \\ \hline
10 & Any nice suggestions of \textit{ENTITY} near the \textit{LOCATION}? I will be going to \textit{LOCATION} the next day. \\ \hline
11 & I just came from \textit{LOCATION}. Someone, please recommend \textit{ENTITY} in the neighborhood of \textit{LOCATION}. \\ \hline
12 & Could anyone propose \textit{ENTITY} not far from the \textit{LOCATION}? I need to leave for \textit{LOCATION} urgently. \\ \hline
13 & We came from \textit{LOCATION} this morning. Suggestions for \textit{ENTITY} close to both \textit{LOCATION} and \textit{LOCATION}? \\ \hline
14 & Any ideas of \textit{ENTITY} between \textit{LOCATION} and \textit{LOCATION}? I would be going to \textit{LOCATION}. Thanks. \\ \hline
15 & We might be staying around \textit{LOCATION}. Please advise \textit{ENTITY} close to \textit{LOCATION} and not far from \textit{LOCATION}. \\ \hline
16 & Could anyone suggest ideas for \textit{ENTITY} close to \textit{LOCATION} and around \textit{LOCATION}? We could be going to \textit{LOCATION} soon. \\ \hline \hline

17 & Any suggestions for \textit{ENTITY} quite far from the \textit{LOCATION}? Thank you very much! \\ \hline
18 & Somebody please suggest \textit{ENTITY} cut off from \textit{LOCATION}. Have a good day! \\ \hline
19 & Does anyone have suggestions for \textit{ENTITY} away from \textit{LOCATION}? Thanks a lot! \\ \hline
20 & Good Afternoon! Any proposals for \textit{ENTITY} not very close to the \textit{LOCATION}? \\ \hline
21 & Suggestions on \textit{ENTITY} far from both \textit{LOCATION} and \textit{LOCATION}? Thank! \\ \hline
22 & Hi! Any idea of \textit{ENTITY} far away from \textit{LOCATION} and \textit{LOCATION}? \\ \hline
23 & Could anyone please propose \textit{ENTITY} not close to \textit{LOCATION} and also far from \textit{LOCATION}? \\ \hline
24 & Does anyone have any suggestions for \textit{ENTITY} far from \textit{LOCATION} and not around \textit{LOCATION}? \\ \hline \hline
25 & Hey! I will be staying at \textit{LOCATION}. Please suggest \textit{ENTITY} cut off from \textit{LOCATION}. \\ \hline
26 & Any pleasant ideas of \textit{ENTITY} far off the \textit{LOCATION}? I might then be visiting \textit{LOCATION}. \\ \hline
27 & I came from \textit{LOCATION} this afternoon. Any proposal for \textit{ENTITY} not close to the \textit{LOCATION}? \\ \hline
28 & Does anyone have a suggestion for \textit{ENTITY} distant from \textit{LOCATION}? By the way, I came from \textit{LOCATION} yesterday. \\ \hline
29 & We will be staying near the \textit{LOCATION}. Suggestions for \textit{ENTITY} far from both \textit{LOCATION} and \textit{LOCATION} will be welcomed. \\ \hline
30 & Any idea of \textit{ENTITY} far away from \textit{LOCATION} and \textit{LOCATION}? I would then be visiting \textit{LOCATION}. \\ \hline
31 & Hi, I will be staying near the \textit{LOCATION}. Could anyone propose \textit{ENTITY} not very close to \textit{LOCATION} and far from \textit{LOCATION}? \\ \hline
32 & Does anyone have suggestions for \textit{ENTITY} far from \textit{LOCATION} and also far from \textit{LOCATION}? I will then be visiting \textit{LOCATION} too. \\ \hline \hline

33 & Any good ideas of \textit{ENTITY} far from \textit{LOCATION} but close to \textit{LOCATION} would be appreciated? Best Regards. \\ \hline
34 & Anyone having ideas of \textit{ENTITY} close to \textit{LOCATION} but far from \textit{LOCATION}? \\ \hline
35 & Someone please advise \textit{ENTITY} far from \textit{LOCATION} but not very far from \textit{LOCATION}. \\ \hline
36 & Suggest \textit{ENTITY} close to \textit{LOCATION} but not in the neighborhood of \textit{LOCATION}. Thank you so much! \\ \hline
37 & Does anyone have good ideas of \textit{ENTITY} far from \textit{LOCATION} but near \textit{LOCATION}? Regards. \\ \hline
38 & Please suggest ideas of \textit{ENTITY} in the neighborhood of \textit{LOCATION} but far from \textit{LOCATION}. \\ \hline
39 & Could anyone advise \textit{ENTITY} far from \textit{LOCATION} but not too far from \textit{LOCATION}? \\ \hline
40 & Any nice ideas of \textit{ENTITY} close to \textit{LOCATION} but not in the neighborhood of \textit{LOCATION}. Thanks! \\ \hline \hline
41 & Tomorrow, I would be coming to stay at \textit{LOCATION}. Anyone having ideas of \textit{ENTITY} close to \textit{LOCATION} but far from \textit{LOCATION}? \\ \hline
42 & Please propose \textit{ENTITY} far from \textit{LOCATION} but not far from \textit{LOCATION}. I will then be exploring \textit{LOCATION}. \\ \hline
43 & I came from \textit{LOCATION} this evening. Any nice ideas for \textit{ENTITY} far from \textit{LOCATION} but close to \textit{LOCATION} would be appreciated?  \\ \hline
44 & Suggest \textit{ENTITY} close to \textit{LOCATION} but not near \textit{LOCATION}. Tomorrow, I will be leaving for \textit{LOCATION}. \\ \hline
45 & Yesterday, I came to stay at \textit{LOCATION}. Any ideas of \textit{ENTITY} close to \textit{LOCATION} but far from \textit{LOCATION}? \\ \hline
46 & Suggestions of \textit{ENTITY} far from \textit{LOCATION} but not very far from \textit{LOCATION}. I will then be moving to \textit{LOCATION}. \\ \hline
47 & I came from \textit{LOCATION} today. Any good ideas for \textit{ENTITY} far from \textit{LOCATION} but near to \textit{LOCATION} would be welcomed?  \\ \hline
48 & Advise \textit{ENTITY} close to \textit{LOCATION} but not close to \textit{LOCATION}. I might be leaving for \textit{LOCATION} soon. \\ \hline
\end{tabular}
\label{tab:templates-table}
\end{table*}

\begin{table*}
\scriptsize
\centering
\caption{List of metonyms for each entity type in the Toy-dataset}
\begin{tabular}{|l|l|}
\hline
\multicolumn{1}{|c|}{\textbf{Entity type}} & \multicolumn{1}{c|}{\textbf{Metonyms}} \\ \hline
R (Restaurant) & \begin{tabular}[c]{@{}l@{}}a restaurant, an eatery, an eating joint, a cafeteria, an outlet, a coffee shop, a fast food place, a lunch counter, \\ a lunch room, a snack bar, a chop house, a steak house, a pizzeria, a coffee shop, a tea house, a bar room\end{tabular} \\ \hline
H (Hotel) & \begin{tabular}[c]{@{}l@{}}a hotel, an inn, a motel, a guest house, a hostel, a boarding house, a lodge, an auberge, a caravansary,\\  a public house,  a tavern, an accomodation, a resort,  a youth hostel, a bunk house, a dormitory, a flop house\end{tabular} \\ \hline
A (Attraction) & \begin{tabular}[c]{@{}l@{}}an attraction, a tourist spot, a tourist attraction, a popular wonder, a sightseeing place, a tourist location, \\ a place of tourist interest, a crowd pleaser,  a scenic spot, a popular landmark, a monument\end{tabular} \\ \hline
\end{tabular}
\label{tab:toy-metonyms-table}
\end{table*}

\subsubsection{Template classes}
We create templates (Table \ref{tab:templates-table})  that can be broadly divided into three different categories based on whether the correct answer entity is expected to be: (1) close to one or more locations [1-16] (2) far from one or more locations [17-32] (3) close to some and far from others (combination) [33-48]. To make the task more reflective of real-world challenges we also randomly insert a {\em distractor} location that does not need to be reasoned. The second-half for each category (i.e. [9-16], [25-32], and [41-48]) consists of templates that have a distractor locative reference. Further, for the close (or far) category, the templates could contain one location ([1-4] + [9-12]) or two locations ([5-8] + [13-16]) that need to be reasoned for close (or far).

\subsubsection{Results} \label{sec:spatial-expts}
We use the following models in our experiments:(i) SPNet (ii) SPNet without (w/o) DRL (iii) BERT-SPNet (iv) BERT-SPNet without (w/o) DRL. Models without DRL use the final hidden states of the Question Encoder and a series of down-projecting feed-forward layers to generate the final score. 

We study our models' performance using $Acc@N$ (N=3,5,30) which requires that any one of the top-N answers be correct, Mean Reciprocal Rank ($MRR$), and the average distance of the top-$3$ ranked answers from the gold-entity $Dist_g$. Table \ref{tab:toy-test-table} summarizes the results test set.

\begin{table*}
\scriptsize
\centering
\caption{Results of the Spatial-reasoning network on the toy-data test set}
\begin{tabular}{|c|l|l|l|l|l|}
\hline
\multicolumn{1}{|c|}{\bf Models} & \textbf{Acc@3} & \textbf{Acc@5} & {\bf Acc@30} & \textbf{MRR} & \textbf{Dg} \\\hline \hline
\multicolumn{1}{|l|}{} & \multicolumn{5}{|c|}{Close to Set X} \\ \hline
SPNet w/o DRL      & 62.60 & 66.00 & 79.00 & 0.608 & 2.88 \\ \hline
SPNet              & 90.20 & {\bf 92.80} & {\bf 97.60} & 0.873 & 0.86 \\ \hline
BERT SPNet w/o DRL & 63.60 & 67.60 & 82.60 & 0.616 & 3.68 \\ \hline
BERT SPNet         & {\bf 91.40} & {\bf 92.80} & 97.20 & {\bf 0.896} & {\bf 0.78} \\ \hline \hline
\multicolumn{1}{|l|}{} & \multicolumn{5}{|c|}{Far from Set X} \\ \hline
SPNet w/o DRL      & 89.00 & 90.80 & 96.40 & 0.858 & 15.24 \\ \hline
SPNet              & {\bf 98.00} & {\bf 98.40} & {\bf 99.20} & 0.975 & 13.88 \\ \hline
BERT SPNet w/o DRL & 90.80 & 92.00 & 95.80 & 0.881 & 15.32 \\ \hline
BERT SPNet         & 97.80 & 98.00 & 98.80 & {\bf 0.978} & {\bf 13.87} \\ \hline \hline
\multicolumn{1}{|l|}{} & \multicolumn{5}{|c|}{Combination} \\ \hline
SPNet w/o DRL      & 23.40 & 28.00 & 50.60 & 0.229 & 9.72  \\ \hline
SPNet              & 52.80 & 60.20 & 82.00 & 0.486 & 3.90  \\ \hline
BERT SPNet w/o DRL & 26.80 & 32.60 & 59.00 & 0.242 & 12.96 \\ \hline
BERT SPNet         & {\bf 59.20} & {\bf 65.80} & {\bf 86.20} & {\bf 0.551} &  {\bf 3.02} \\ \hline \hline
\multicolumn{1}{|l|}{} & \multicolumn{5}{|c|}{Aggregate} \\ \hline
SPNet w/o DRL      & 58.33 & 61.60 & 75.33 & 0.565 & 9.28  \\ \hline
SPNet              & 80.33 & 83.80 & 92.93 & 0.778 & 6.21  \\ \hline
BERT SPNet w/o DRL & 60.40 & 64.07 & 79.13 & 0.579 & 10.65 \\ \hline
BERT SPNet         & {\bf 82.80} & {\bf 85.53} & {\bf 94.07} & {\bf 0.808} & {\bf 5.89}  \\ \hline
\end{tabular}
\label{tab:toy-test-table}
\end{table*}

\subsubsection{Error Analysis}
Tables \ref{tab:toy-error-table1} and \ref{tab:toy-error-table2} show the effect of candidate search space and the number of location mentions in the question on the performance of the SPNet Model.

\begin{table}[H]
\scriptsize
\hspace{-2ex}
\caption{Performance of SPNet decreases with increase in universe size.}
\begin{tabular}{|l|l|l|l|l|l|l|}
\hline
\multicolumn{1}{|c|}{\textbf{}} & \multicolumn{2}{c|}{\textbf{Correctly Answered}} & \multicolumn{2}{c|}{\textbf{Incorrectly Answered}} \\ \hline
\multicolumn{1}{|c|}{\textbf{Search Space size}} & \multicolumn{1}{c|}{\textbf{Questions}} & \multicolumn{1}{c|}{\textbf{Percentage}} & \multicolumn{1}{c|}{\textbf{Questions}} & \multicolumn{1}{c|}{\textbf{Percentage}} \\ \hline
0-200      & 318 & 26.39\% & 42 & 14.24\%  \\  \hline
200-500    & 417 & 34.61\% & 83 & 28.13\% \\ \hline
500-1000   & 221 & 18.34\% & 53 & 17.97\% \\ \hline
1000-5000  & 178 & 14.77\% & 82 & 27.80\% \\ \hline
5000-20000 & 71  & 5.89\%  & 35 & 11.86\% \\ \hline
\end{tabular}
\label{tab:toy-error-table1}
\end{table}

\begin{table}[H]
\scriptsize
\centering
\caption{Performance of SPNet decreases with increase in the number of location mentions in the question.}
\begin{tabular}{|c|l|l|l|l|l|l|}
\hline
\multicolumn{1}{|c|}{\textbf{}} & \multicolumn{2}{c|}{\textbf{Correctly Answered}} & \multicolumn{2}{c|}{\textbf{Incorrectly Answered}} \\ \hline
\multicolumn{1}{|c|}{\textbf{\# Location-Mentions}} & \multicolumn{1}{c|}{\textbf{Questions}} & \multicolumn{1}{c|}{\textbf{Percentage}} & \multicolumn{1}{c|}{\textbf{Questions}} & \multicolumn{1}{c|}{\textbf{Percentage}} \\ \hline
1 & 233 & 19.34\%  & 0  & 0.00\%  \\  \hline
2 & 671 & 55.69\% & 126 & 42.71\% \\ \hline
3 & 301 & 24.98\% & 169 & 57.29\% \\ \hline
\end{tabular}
\label{tab:toy-error-table2}
\end{table}

\subsection{Location Tagger} \label{sec:loctagger}

In order to get mentions of locations in questions, we manually label a set of $425$ questions from the training set for location mentions. We then use a BERT-BiLSTM CRF \cite{MSEQ} based tagger trained on this set to label locations. Table \ref{tab:location-tagger-results} describes the performance of the tagger on an unseen set of $75$ questions. 


\begin{table}[H]
\scriptsize
\centering
\caption{Performance of the BERT-BiLSTM CRF for tagging locations on a small set of 75 unseen questions.}
\begin{tabular}{|l|l|l|l|}
\hline
 & \textbf{Precision} & \textbf{Recall} & \textbf{F1} \\ \hline
\textbf{Micro Average} & 87.59 & 87.56 & 87.58 \\ \hline
\textbf{Macro Average} & 88.24 & 87.83 & 88.03 \\ \hline
\end{tabular}%
\label{tab:location-tagger-results}
\end{table}


\subsection{Spatio-textual Reasoning Network} \label{sec:spatio-textual-expts}
The Spatio-Textual Reasoning Network consists of three components (i) Spatial Reasoner (ii) Textual Reasoner (iii) Joint Scoring Layer.   

\noindent{\bf Training: } We train the joint model using max-margin loss teaching the network to score the correct-answer higher than a negatively sampled candidate entity. Model parameters are described in the next section.

\subsubsection{Results}
Similar to \citet{TourQue} we also experiment on this dataset by employing a neural method to reduce the search space \cite{DUET2} before using the $\crq$ textual-reasoner to re-rank only the top-30 selected candidates (pipeline referred to as $\csr$). Unlike $\crq$, which uses two levels of attention between question and review sentences to score candidate entities $\cs$ does not reason deeply over the text. It compares elements of a question with different parts of a review document to aggregate relevance for scoring. Local and distributed representations are used to capture lexical and semantic features.

We report some experiments using this model referred to as $\cs$ and compare it with $\csr$ and spatio-textual $\csr$. As can be seen re-ranking with SD or SPNet does not help the system. An interesting direction of future work could thus, be to augment general-purpose neural-IR methods such as Duet used by $\cs$ with spatial-reasoning. Another interesting approach could be to extend ideas from existing Graph-neural network based approaches, such as NumNet \cite{NumNet}. Each entity could be viewed as a node in a graph for reasoning but we note that methods will need to be made more scalable for them to be useful. The entity space (and thus nodes in the graph) would run into thousands of nodes per question making current message-passing based inference methods prohibitively expensive. 

\begin{table}
\scriptsize
\center
\caption{Comparison of re-ranking models operating on a reduced search space returned by $\cs$ on Location Questions (ii) Comparison with current state-of-the-art $\csr$ on the full task.}
\begin{tabular}{|c|l|l|l|l|l|}
\hline
\multicolumn{1}{|l|}{}                                          & \multicolumn{5}{|c|}{Location Questions}                                \\ \hline
\multicolumn{1}{|c|}{\bf Models}                                          & \textbf{Acc@3} & \textbf{Acc@5} & {\bf Acc@30} & \textbf{MRR} & \textbf{Dg} \\\hline
$\cs$                                                          &      15.84          &     20.26        &  {\bf 51.47}    &   0.149    &      2.61      \\ \hline
$\cs$ $\rightarrow$ SD        &   11.34             &   17.26     &   {\bf 51.47}     &      0.118        & {\bf 2.18}             \\ \hline
$\cs$ $\rightarrow$ LocNet     &      8.38          &  13.72     &  {\bf 51.47}       & 0.097                &    2.27         \\ \hline 
$\csr$ & 19.89 & 26.43 & {\bf 51.47} & 0.168 & 2.70 \\ \hline
\begin{tabular}[c]{@{}c@{}}Spatio-textual\\ CSRQA\end{tabular} 
& {\bf 21.36} & {\bf 28.36} & {\bf 51.47} & {\bf 0.183} & 2.27 \\ \hline
\multicolumn{1}{|l|}{}                                          & \multicolumn{5}{|c|}{All Questions}                                \\ \hline

$\csr$ & 21.45 & 28.21 & {\bf 52.65} & 0.186 & 2.47 \\ \hline
\begin{tabular}[c]{@{}c@{}}Spatio-textual\\ CSRQA\end{tabular} 
& {\bf 22.41} & {\bf 28.99} & {\bf 52.65} & {\bf 0.193} & {\bf 2.32}\\ \hline

\end{tabular}
\label{tab:csr-loc-results}
\end{table}

\subsection{Model settings} \label{sec:model-params}

\subsubsection{Experiments on Toy Dataset}




The hyperparameters for the best performing configurations of all models were identified through manual testing on the validation set  (Table \ref{tab:toy-hyper-table}). The models were trained on a 2x NVIDIA K40 (12GB, 2880 CUDA cores) GPU on a shared cluster.

\begin{table}
\scriptsize
\centering
\caption{Hyperparameter settings for experiments on the toy-dataset} 
\begin{tabular}{|l|l|}
\hline
\multicolumn{1}{|c|}{\textbf{Hyperparameter}} & \multicolumn{1}{c|}{\textbf{Value}} \\ \hline
Negative samples & 40 \\ \hline
Batch size & 20 \\ \hline
Optimizer & Adam \\ \hline
Loss & MarginRankingLoss \\ \hline
Margin & 0.5 \\ \hline
Max no. of epochs & 15 \\ \hline 
GRU Input dimension & 131 \\ \hline
GRU Output dimension & 32 \\ \hline
DRL Block Layer 1 & 64 (Input) 64 (Output) \\ \hline
DRL Block Layer 2 & 64 (Input) 64 (Output) \\ \hline
DRL Block Layer 3 & 64 (Input) 64 (Output) \\ \hline
DRL Block Layer 4 & 64 (Input) 1 (Output) \\ \hline

\end{tabular}
\label{tab:toy-hyper-table}
\end{table}

The BERT models were trained with a learning rate of 0.0002 whereas the non-BERT models with a learning rate of 0.001. 





\balance

\subsubsection{Spatio-textual Reasoning Network}




The hyperparameters for the best performing configuration were identified through manual testing on the validation set (Table \ref{tab:main-hyper-table}). The  Spatio Textual Reasoner was trained on 4 K-80 GPUs on a shared cluster.

\begin{table}
\scriptsize
\centering
\caption{Hyperparameters used for experiments on the end-task} 
\begin{tabular}{|l|l|}
\hline
\multicolumn{1}{|c|}{\textbf{Hyperparameter}} & \multicolumn{1}{c|}{\textbf{Value}} \\ \hline
Word embeddings size & 128 \\ \hline
Dropout & 0.2 \\ \hline
Optimizer & Adam \\ \hline
Loss & Hinge Loss \\ \hline
Margin & 1.0 \\ \hline
Batch Size & 200 \\ \hline
$\locnet$ GRU input dimension & 131 \\ \hline
$\locnet$ GRU output dimension & 256 \\ \hline
Textual GRU input dimension & 128 \\ \hline
Textual GRU output dimension & 256 \\ \hline
DRL Block Layer 1 & 512 (Input) 256 (Output)\\ \hline
DRL Block Layer 2 & 256 (Input) 256 (Output)\\ \hline
DRL Block Layer 3 & 256 (Input) 128 (Output)\\ \hline
DRL Block Layer 4 & 128 (Input) 128 (Output)\\ \hline
DRL Block Layer 5 & 128 (Input) 50 (Output)\\ \hline
DRL Block Layer 6 & 50 (Input) 10 (Output)\\ \hline
DRL Block Layer 7 & 10 (Input) 1 (Output)\\ \hline
$\alpha$,$\beta$ FF Linear Layer 1 & 256 (Input) 50 (Output) \\ \hline
$\alpha$,$\beta$ FF Linear Layer 2 & 50 (Input) 50 (Output) \\ \hline
$\alpha$,$\beta$ FF Linear Layer 3 & 50 (Input) 10 (Output) \\ \hline
$\alpha$,$\beta$ FF Linear Layer 4 & 10 (Input) 10 (Output) \\ \hline
$\alpha$,$\beta$ FF Linear Layer 5 & 10 (Input) 10 (Output) \\ \hline
$\alpha$,$\beta$ FF Linear Layer 6 & 10 (Input) 2 (Output) \\ \hline
\end{tabular}
\label{tab:main-hyper-table}
\end{table}

\end{document}